\documentclass[]{article} 

\usepackage{proceed2e} % UAI 2014 style

\usepackage{amsmath,amsfonts,amssymb}
\usepackage{mathrsfs}
\usepackage{amsbsy}
\usepackage{natbib}
\usepackage{url}
\usepackage{caption}
\usepackage{subcaption}
\usepackage{graphicx}
\usepackage[compact]{titlesec}
\usepackage{hyperref}
\usepackage{epstopdf}
\usepackage{times}
\usepackage{bm}
\usepackage{latexsym}
\usepackage{amsthm}

\global\long\def\dataDim{p}
\global\long\def\latentDim{q}

\global\long\def\numData{n}

\global\long\def\dataScalar{y}

\global\long\def\dataMatrix{\mathbf{\MakeUppercase{\dataScalar}}}

\global\long\def\latentScalar{x}
\global\long\def\latentMatrix{\mathbf{\MakeUppercase{\latentScalar}}}
\global\long\def\latentVector{\mathbf{\latentScalar}}

\global\long\def\numData{n}

\global\long\def\dataDim{p}

\usepackage{color}
\usepackage{verbatim}

\definecolor{brown}{rgb}{0.9,0.59,0.078}
\definecolor{ironsulf}{rgb}{0,0.7,.5}
\definecolor{lightpurple}{rgb}{0.156,0,0.245}

\ifdefined\blackBackground
\definecolor{colorOne}{rgb}{0, 1, 1}
\definecolor{colorTwo}{rgb}{1, 0, 1}
\definecolor{colorThree}{rgb}{1, 1, 0}
\definecolor{colorTwoThree}{rgb}{1, 0, 0}
\definecolor{colorOneThree}{rgb}{0, 1, 0}
\definecolor{colorOneTwo}{rgb}{0, 0, 1}
\else
\definecolor{colorOne}{rgb}{1, 0, 0}
\definecolor{colorTwo}{rgb}{0, 1, 0}
\definecolor{colorThree}{rgb}{0, 0, 1}
\definecolor{colorTwoThree}{rgb}{0, 1, 1}
\definecolor{colorOneThree}{rgb}{1, 0, 1}
\definecolor{colorOneTwo}{rgb}{1, 1, 0}
\fi

\ifdefined\blackBackground

\else

\fi

\global\long\def\bfmu{\boldsymbol{\mu}}

\global\long\def\bfSigma{\boldsymbol{\Sigma}}

\global\long\def\bfa{\mathbf{a}}
\global\long\def\bfb{\mathbf{b}}

\global\long\def\bff{\mathbf{f}}

\global\long\def\bfw{\mathbf{w}}
\global\long\def\bfx{\mathbf{x}}
\global\long\def\bfy{\mathbf{y}}

\global\long\def\cov{\text{cov}}

\global\long\def\bfG{\mathbf{G}}

\global\long\def\bfI{\mathbf{I}}

\global\long\def\bfK{\mathbf{K}}

\global\long\def\bfW{\mathbf{W}}
\global\long\def\bfX{\mathbf{X}}
\global\long\def\bfY{\mathbf{Y}}

\global\long\def\cut#1{}

\global\long\def\detail#1{}

\newcommand{\kss}{\bfK_{\bf{*},\bf{*}} }
\newcommand{\kxxt}{\tilde{\bfK}_{\bf{x},\bf{x}} } 
\newcommand{\kxst}{\tilde{\bfK}_{\bf{*},\bf{x}} } 
\newcommand{\ksxt}{\tilde{\bfK}_{\bf{x},\bf{*}} } 
\newcommand{\ksst}{\tilde{\bfK}_{\bf{*},\bf{*}} }
\newcommand{\vect}[1]{\operatorname{vec}#1}
\theoremstyle{definition}               %roman style
\newtheorem*{defin}{Definition} %
\renewcommand\numData{N}  % number of data inputs

\global\long\def\bfJ{\mathbf{J}}

\title{Metrics for Probabilistic Geometries} 

\author{} % LEAVE BLANK FOR ORIGINAL SUBMISSION.
\author{ {\bf Alessandra Tosi} \\  
Dept. of Computer Science \\ 
Universitat Polit\`ecnica \\
de Catalunya \\
Barcelona, Spain \\ 
\And 
{\bf S{\o}ren Hauberg}  \\ 
DTU Compute \\
Technical University \\ 
of Denmark \\
Denmark  \\           
\And 
{\bf Alfredo Vellido}   \\ 
Dept. of Computer Science \\ 
Universitat Polit\`ecnica \\
de Catalunya \\
Barcelona, Spain \\ 
\And 
{\bf Neil D. Lawrence\thanks{ ~Also at Sheffield Institute for Translational Neuroscience, SITraN. Sheffield, UK}}   \\ 
Dept. of Computer Science \\
The University \\
of Sheffield \\
Sheffield, UK
} 
\begin{document} 
 
\maketitle 
 
\begin{abstract} 

We investigate the geometrical structure of probabilistic generative dimensionality reduction models using  the tools of Riemannian geometry. We explicitly define a distribution over the natural metric given by the models. We provide the necessary algorithms to compute expected metric tensors where the distribution over mappings is given by a Gaussian process. We treat the corresponding latent variable model as a Riemannian manifold and we use the expectation of the metric under the Gaussian process prior to define interpolating paths and measure distance between latent points. We show how distances that respect the expected metric lead to more appropriate generation of new data.

\end{abstract} 
 
\section{MOTIVATION} \label{sec:intro}

One way of representing a high dimensional data set is to relate it to a lower dimensional set of \emph{latent variables} through a set of (potentially nonlinear) functions. If the $i$th data point and the $j$th feature is represented by $y_{i, j}$, it might be related to a $q$ dimensional vector of latent variables $\mathbf{x}_{i, :}$ as
\[
y_{i,j} = f_j(\mathbf{x}_{i, :}) + \boldsymbol{\epsilon}_{i},
\]
where $f_j(\cdot)$ is a nonlinear function mapping to the $j$th
feature of the data set and $\boldsymbol{\epsilon}_{i}$ is a noise
corruption of the underlying function. A manifold derived from a
finite data set can never be precisely determined across the entire
input range of $\mathbf{x}$. We consider posterior distributions
defined over $f_j(\cdot)$ and we focus on the uncertainty defined over
the local metric of the manifold itself. This allows us to define
distances that are based on metrics that take account of the
uncertainty with which the manifold is defined. We use these metrics
to define distances between points in the latent space that respect
these metrics.

\begin{figure}[t]
  \includegraphics[trim=0 0 0 0,clip,scale=0.4]{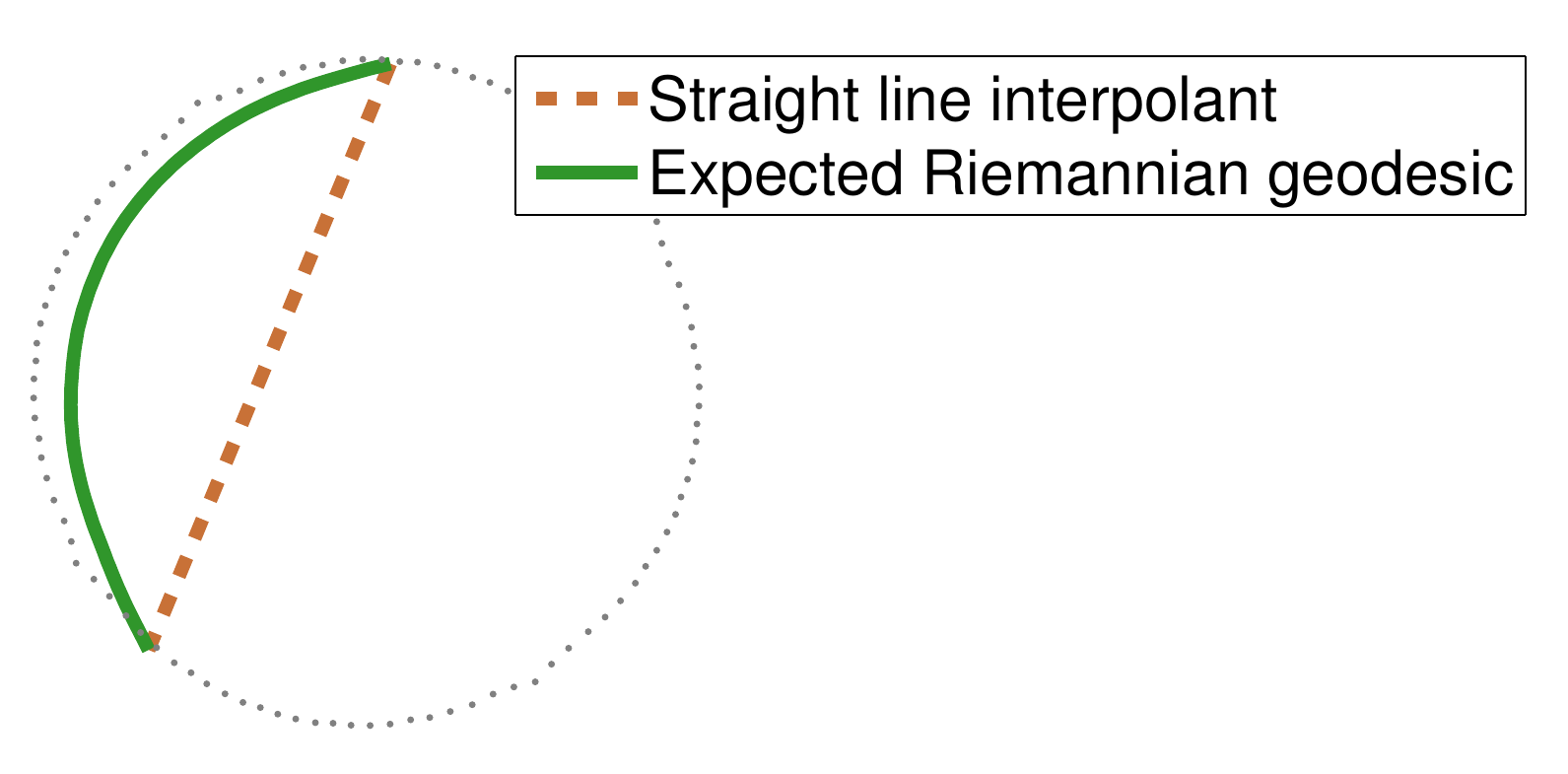}
  \raisebox{5mm}{\makebox[0pt]{\hspace{-1cm}
    \fbox{\includegraphics[width=1cm]{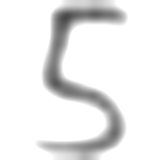}}
    \fbox{\includegraphics[width=1cm]{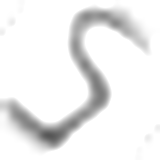}}
    \fbox{\includegraphics[width=1cm]{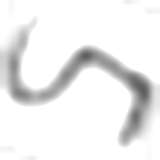}}}}
\vskip .5pc 
\caption{The latent space from a GP-LVM that was trained over a
  dataset of artificially rotated digits. Black  dots represent the latent points.
   The dashed brown line show the commonly used straight-line interpolant, and the green curve is
  the suggested expected Riemannian geodesic. This figure is best
  viewed in colour.}
\label{fig:introduction}
\end{figure}

When the mappings $f_j(\cdot)$ are nonlinear, the latent variable model (LVM) can potentially
capture non-linearities on the data and thereby provide an even lower
dimensional representation as well as a more useful view of the
data. While this line of thinking is popular, it is not without its
practical issues. As an illustrative example,
Fig.~\ref{fig:introduction} shows the latent representation of a set of
artificially rotated images obtained through a Gaussian process
latent variable model (GP-LVM). It is clear from the display that the latent
representation captures the underlying periodic structure of the
process which generated the data (a rotation). If we want to analyse
the data in the latent space, e.g. by interpolating latent points, our
current tools are insufficient. As can be seen, fitting a straight
line in the latent space between the two-points leads to a solution that does not interpolate well in the data space: the interpolant goes through
regions where the data does not reside, regions where the actual functions, $f_j(\cdot)$, cannot be well determined.

This observation raises several related questions about the choice of
interpolant: 1) what is the natural choice of interpolant in the
latent space? And, 2) if the natural interpolant is not a straight
line, are Euclidean distances still meaningful? We answer these
questions for the GP-LVM, though our approach is applicable to other
generative models as well. We consider here a metric which reflects
the intrinsic properties of the original data and recovers some
information loss due to the nonlinear mapping performed by the
model. We find that for smooth LVMs the metric from the observation
space can be brought back to the latent space in the form of a random
Riemannian metric. We then provide algorithms for computing distances
and shortest paths (geodesics) under the expected Riemannian
metric. With this the natural interpolant becomes a curve, which
follows the trend of the data.

\paragraph{Overview}
In Section~\ref{sec:riem} we introduce the concepts of Riemannian
geometry, the tool on which we rely on later on in the paper. Section
\ref{sec:prob_dim_redux} provides an overview of the state of the art
in probabilistic dimensionality reduction, introducing the class of
models to which the proposed methodology can be extended.
In Section~\ref{sec:gp-lvm} we use the probabilistic nature of the
generative LVMs to explicitly provide distributions over the metric
tensor; first, we provide the general expressions, then we specialise
these to the GP-LVM as an example. Finally, we show how to compute
shortest paths (geodesics) over the latent space.
Experimental results are provided in Section~\ref{sec:exp}, and
the paper is concluded with a discussion in Section~\ref{sec:discussion}.

\section{CONCEPTS OF RIEMANNIAN GEOMETRY} \label{sec:riem} 

We study latent variable models (LVMs) as embeddings of uncertain surfaces (or
manifolds) into the observation space.  From a machine learning point
of view, we can interpret this embedded manifold as the underlying
support of the data distribution.  To this end, we review the basic
ideas of differential geometry, which study surfaces through local
linear models.
  
\begin{figure}
    \includegraphics[width=0.45\textwidth]{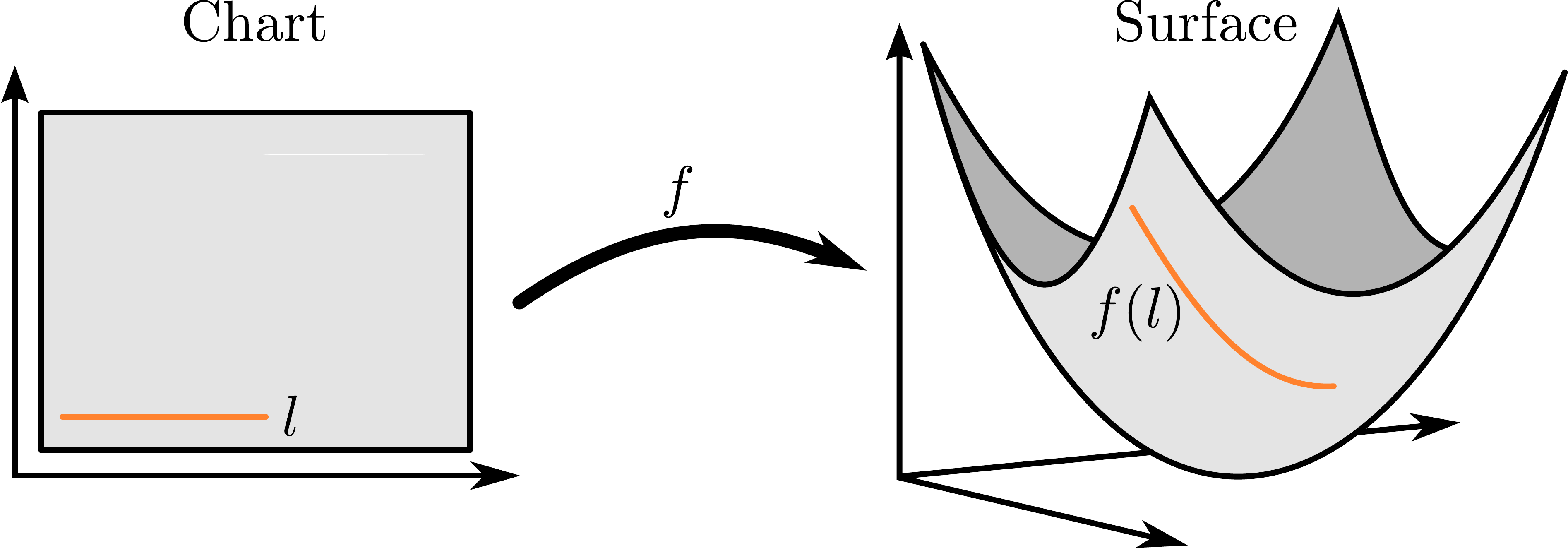}
    \vskip .5pc 
    \caption{An illustration of the standard surface model; $f$ maps the
      chart into the embedding space.}
    \label{fig:surface_drawing}
  \end{figure}

Gauss' study \citeyearpar{gauss:surfaces:1827} of curved surfaces is among
the first examples of (deterministic) LVMs.  He noted that a
$\latentDim$-dimensional surface embedded in a $\dataDim$-dimensional
Euclidean space\footnote{Historically, Gauss considered the case of two-dimensional
  surfaces embedded in $\mathbb{R}^3$, while the extension to higher dimensional
  \emph{manifolds} is due to Bernhard Riemann.} is well-described through a mapping $f:
\mathbb{R}^\latentDim \rightarrow \mathbb{R}^\dataDim$.  The
$\latentDim$-dimensional representation of the surface is known as the
\emph{chart} (in machine learning terminology, this corresponds to the
\emph{latent space}).  In general, the mapping $f$ between the chart
and the embedding space is not \emph{isometric}, e.g.\ the Euclidean
length of a straight line $l$ in the chart does not match the length
of the embedded curve $f(l)$ as measured in the embedding
space. Intuitively, the chart provides a distorted view of the surface
(see Fig.~\ref{fig:surface_drawing} for an illustration).  To rectify
this view, Gauss noted that the length of a curve is
  \begin{align}
    \mathrm{Length}\left( f(l) \right)
      &= \int_0^1 \left\| \frac{\partial f(l(t))}{\partial t} \right\| dt
      &= \int_0^1 \left\| \bfJ \frac{\partial l(t)}{\partial t} \right\| dt,
    \label{eq:len1}
  \end{align}
  where $\bfJ$ denotes the Jacobian of $f$, i.e.
  \begin{equation}
  \left[ \bfJ \right]_{i, j} = \frac{\partial f_i}{\partial l_j}.
  \label{eq:jacobian}
  \end{equation}
  Measurements on the surface can, thus, be computed
  in the chart locally, and integrated to provide global measures. This gives rise to the definition
  of a \emph{local} inner product, known as a \emph{Riemannian metric}.
  
  \begin{defin}[Riemannian Metric]
    A Riemannian metric $\bfG$ on a manifold $\mathcal{M}$ is a
    symmetric and positive definite matrix which defines a smoothly
    varying inner product
    \begin{align}\label{eq:tensor}
      \langle \bfa, \bfb \rangle_x &= \bfa^\top \bfJ^\top \bfJ \bfb = \bfa^\top \bfG(x)\bfb
    \end{align}
    in the tangent space $T_{x}\mathcal{M}$, for each point $x \in \mathcal{M}$ and
    $\bfa, \bfb \in T_{x}\mathcal{M}$. The matrix $\bfG$ is called the \emph{metric tensor}.
  \end{defin}
  \paragraph{Remark} The Riemannian metric need not be restricted to
  $\bfG = \bfJ^\top \bfJ$ and can be any smoothly changing symmetric
  positive definite matrix \citep{carmo:1992}.  We restrict ourselves
  to the more simple definition as it suffices for our purposes, but
  note that the more general approach has been used in machine
  learning, e.g.\ in \emph{metric learning} \citep{hauberg:nips:2012}
  and \emph{information geometry} \citep{amari:book:2000}.
  
  From this definition, Eq.~\ref{eq:len1} reduces to
  \begin{align}
    \mathrm{Length}\left( \gamma \right)
      &= \int_0^1 \sqrt{ \langle \gamma'(t), \gamma'(t) \rangle_{\gamma(t)} } dt
    \label{eq:curve_length}
  \end{align}
  for a general curve $\gamma: [0, 1] \rightarrow \mathbb{R}^\dataDim$.
  
  \begin{defin}[Geodesic curve]
    Given two points  $\bfx_1, \bfx_2 \in \mathcal{M}$, a \emph{geodesic} is a
    length-minimising curve connecting the points
    \begin{align}
      \gamma_g = \arg\min_{\gamma} \mathrm{Length}(\gamma), \quad \gamma(0) = \bfx_1, \gamma(1) = \bfx_2.
    \end{align}
  \end{defin}
  It can be shown \citep{carmo:1992} that geodesics satisfy the following second
  order ordinary differential equation
  \begin{align}\label{eq:geodesic_ode}
    \gamma'' &= 
      -\frac{1}{2} \bfG^{-1} \left[ \frac{\partial \vect{\bfG}}{\partial \gamma} \right]^\top
      (\gamma' \otimes \gamma'),
  \end{align}
  where $\vect{\bfG}$ stacks the columns of $\bfG$ and $\otimes$
  denotes the Kronecker product. The Picard-Lindel{\"o}f theorem
  \citep{tenenbaum} then implies that geodesics exist and are locally
  unique given a starting point and an initial velocity.

\section{PROBABILISTIC DIMENSIONALITY REDUCTION}\label{sec:prob_dim_redux}

Nonlinear dimensionality reduction methods  \citep{LeeVerley:2007} provide a flexible data
representation which can provide a more faithful model of the observed
multivariate datasets than the linear ones. One approach
is to perform probabilistic nonlinear dimensionality reduction
defining a model that introduces a set of unobserved (or latent)
variables $\latentMatrix$ that can be related to the observed ones
$\bfY$ in order to define a joint distribution over both. These
models are known as latent variable models (LVMs). The latent space is
dominated by a prior distribution $p(\latentMatrix)$ which induces a
distribution over $\bfY$ under the assumption of a probabilistic
mapping
\begin{equation}
y_{i,j} = f_j(\bfx_i) + \epsilon_i,
\label{eq:lvm}
\end{equation}
where $\bfx_i$ is the latent point associated with the $i^{th}$
observation $\bfy_i$, $j$ is the index of the features of $\bfY$, and
$\epsilon_i$ is a noise term, accounts for both noise in the data as
well as for inaccuracies in the model. The noise is typically chosen
as Gaussian distributed $\epsilon \sim \mathcal{N}(0,\beta^{-1})$,
where $\beta$ is the precision.

One of the advantages of this approach is that it accommodates
dimensionality reduction in an intuitive manner, if we assume that the
dimensionality of the latent space is significantly lower than that of
the observation space. In this case, the reduced dimensionality
provides us with both implicit regularisation and a low-dimensional
representation of the data, which can be used for visualisation (and,
therefore, for data exploration \citep{Vellido:2011:esann}) if the dimension is low enough.

If the mapping $f = W$ is taken to be linear:
\begin{equation}
y_{i,j} = \bfw_j^\top \bfx_i + \epsilon_i,
\end{equation}
and the prior  $p(\latentMatrix)$ to be Gaussian, this model is known as probabilistic principal component analysis \citep{Tipping:pca97}. The conditional  probability of the data given the latent space can be written as 
\begin{equation}
p(\bfy_i \mid \bfx_i, \bfW, \beta) = \mathcal{N}(\bfy_i \mid \bfW \bfx_i, \beta^{-1}\bfI).
\end{equation}
With a further assumption of independence across data points, the marginal likelihood of the data is
\begin{equation}
p(\bfY) = \int \prod_{i=1}^{N} p(\bfy_i \mid \bfx_i, \bfW, \beta) p(\bfx_i) d\latentMatrix .
\end{equation}

In general, this approach can be applied to both linear and nonlinear dimensionality reduction models, leading to the definition of, for instance, Factor Analysis \citep{Bartholomew:1987}, Generative Topographic Mapping (GTM) \citep{Bishop:gtmncomp98}, or Gaussian Process-LVM (GP-LVM) \citep{Lawrence:pnpca05} to name a few.

One example that generalises from the linear case to the nonlinear one is the GTM, in which the noise model is taken to be a linear combinations of a set of $M$ basis functions
\begin{equation}
y_{i,j} = \sum_{m=i}^M \bfw_j^\top \phi_m(\bfx_i) + \epsilon_i.
\end{equation}
This model can be seen as a mixture of distributions (usually Gaussian radial basis distributions) whose centres are constrained to lay on an intrinsically low-dimensional space. These centres can be interpreted as data prototypes or cluster centroids that can be further agglomerated in a full blown clustering procedure. In this manner, GTM mixes the functionalities of Self-Organising Maps and mixture models by providing both data visualisation over the latent space and data clustering \citep{Olier:2008:NN}. If the prior over the latent space is chosen to be Gaussian, this model leads, in a similar way of probabilistic PCA, to a Gaussian conditional distribution of the data
\begin{equation}
p(\bfy_i \mid \bfx_i, \bfW, \beta) = \mathcal{N}\left( \bfy_i \left\vert \sum_{m=i}^M \right. \bfw_j^\top \phi_m(\bfx_i), \beta^{-1}\bfI \right).
\label{eq:gtm}
\end{equation}

In the classic approach the latent variables are marginalised and the parameters are optimised by maximising the model likelihood. An alternative (and equivalent) approach proposes to marginalise the parameters and optimise the latent variables, leading to Gaussian Process Latent Variables Model (GP-LVM). %\citep{Lawrence:pnpca05}. 

In terms of applications, \citet{grochow:siggraph:2004} animate human poses using \emph{style-based inverse kinematics}
based on a GP-LVM model. The animation is performed under a prior towards
small Euclidean motions in the latent space, i.e.\ under the same assumptions as those
leading to a straight-line interpolant. As the Euclidean metric does not match that of the
observation space, this prior is difficult to interpret.
In a related application, \citet{urtasun:iccv:2005} track the pose of a person in a video
sequence with a similar prior and, hence, similar considerations hold.
Recently, \citet{gonczarek:2014} track human poses in images under a Brownian motion
prior in the latent space. Again, this relies on a meaningful metric in the latent space.
In all of the above application, it is beneficial if the metric in the latent space is
related to that of the observation metric.

\section{METRICS FOR PROBABILISTIC LVMs} \label{sec:gp-lvm} 

The common approach to estimate local metrics relies on assumptions over the neighbourhoods
defined in the observed space (see e.g.\ \citep{hastie:tpami96, ramanan:tpami11}).
This might be less efficient in presence of high dimensional noise, because the induced distances may not be reliable. One way to deal with this problem is to define a noise model \eqref{eq:lvm} and to assume a global belief over the geometry of the data. This way, the resulting models have the advantage of providing an intrinsic local metric which is able to deal with noise.

In this paper we only consider smooth generative models for manifold learning. This contrasts with prior approaches such as \citep{bregler:nips:1994, tenenbaum:nips:1997, tenenbaum:science:2000} that use metrics which vary discretely across the space (see also \citep{Lawrence:unifying12} for relations to Gaussian models). 

We define here the local metric tensor for generative LVMs. We then illustrate the specific case of GP-LVM, providing an algorithm to compute shortest path.

\subsection{THE DISTRIBUTION OF THE NATURAL METRIC} \label{subsec:local_metric}

When the mapping $f$ in Eq.~\ref{eq:lvm} is differentiable, it can be interpreted as the mapping between the \emph{chart} (or \emph{latent space}) and the embedding space (c.f.\ Section~\ref{sec:riem}). Then it is possible to explicitly compute the natural Riemannian metric of the given model. 

Let $\bfJ$ be the Jacobian (as in Eq.~\ref{eq:jacobian}), then the tensor 
$$\bfG = \bfJ^\top\bfJ$$ 
defines a local inner product structure over the latent space according to Eq.~\ref{eq:tensor}.

In the case of LVMs where the conditional probability over the Jacobian follow a Gaussian distribution, this naturally induces a distribution over the local metric tensor $\bfG$. Assuming independent rows of $\bfJ$
\begin{equation}
p(\bfJ \mid \bfX, \beta) = \prod_{j=1}^\dataDim \mathcal{N}(\bfJ_{j,:} \mid \bm{\mu}_{J_{j,:}},\bfSigma_J),
\end{equation}
the resulting random variable follow a non-central Wishart
distribution \citep{anderson:annals:1946}:
\begin{equation}
\bfG = \mathcal{W}_{\latentDim}( \dataDim, \bfSigma_J, \mathbb{E}[\bfJ^\top]\mathbb{E}\left[\bfJ\right]),
\label{eq:wishart}
\end{equation}
where $\dataDim$ represents the number of degrees of freedom; the quantity $\bfSigma_J^{-1}\mathbb{E}[\bfJ^\top]\mathbb{E}\left[\bfJ\right]$ is know as the non-centrality matrix and it is equal to zero in the central Wishart distribution. The Wishart distribution is a multivariate generalisation of the Gamma distribution. 

\subsection{GP-LVM LOCAL METRIC}

A Gaussian Process (GP) is used to describe distributions over functions and it is defined as a collection of random variables, any finite number of which have a joint Gaussian distribution \citep{Rasmussen:book06}. Given a vector $\latentVector \in \mathbb{R}^{\latentDim}$, a GP determined by its mean function and its covariance function is denoted $f(\bfx) \sim \mathcal{GP}(m(\bfx),k(\bfx,\bfx'))$. From this, it is possible to generate a random vector $\bff$ which is Gaussian distributed with covariance matrix given by $(\bfK)_{i,j} = k(\bfx_i,\bfx_j)$.

Gaussian Processes have been used in probabilistic nonlinear dimensionality reduction to define a prior distribution over the mapping $f$ (in Eq. \ref{eq:lvm}), leading to the formulation of GP-LVM. This way, the likelihood of the data $\bfY$ given $\latentMatrix$ is computed by marginalising the mapping and optimising the latent variables:
\begin{equation}
p(\dataMatrix\!\!\mid\!\latentMatrix,\!\beta)\!=\!\!\prod_{j=1}^\dataDim\!\mathcal{N}(\bfy_{:,j}\!\mid\!\bm{0},\bfK\! +\!\beta^{-1}\bfI)\!=\!\!\prod_{j=1}^\dataDim\!\mathcal{N}(\bfy_{:,j}\!\!\mid\!\bm{0},\tilde{\bfK}).
\end{equation}

To follow the notation introduced in Section~\ref{sec:prob_dim_redux}, the noise model is defined by
\begin{equation}
y_{i,j} = \tilde{\bfK}_{(\bfx_i,\bfX)} \tilde{\bfK} \bfY_{:,j}  + \epsilon_i,
\end{equation}

Due to the linear nature of the differential operator, the derivative of a Gaussian process is again a Gaussian process (\citep{Rasmussen:book06} \S 9.4), as long as the covariance function is differentiable. This property allows inference and predictions about derivatives of a Gaussian Process, therefore the Jacobian $\bfJ$ of the GP-LVM mapping can be computed over continuum for every latent point $\bfx_*$ and we denote with $\frac{\partial \bfy_*}{ \partial x^{(i)}}$ the partial derivative of $\bfy(x_*)$  with respect to the $i^{th}$  component in the latent space. We call $\bfJ^\top_* \!= \frac{\partial \bfy_*}{\partial \bfx} =\! \left[ \frac{\partial \bfy_*}{\partial x^{(1)}}; \cdots; \frac{\partial \bfy_*}{\partial x^{(\latentDim)}} \right]$, where $\frac{\partial \bfy_*}{\partial \bfx}$ is a $\latentDim \times \dataDim$ matrix whose columns are multivariate normal distributions. We now consider the jointly Gaussian random variables
\begin{equation}
\left[ \begin{array}{c} \dataMatrix \\ \frac{\partial \bfy_*}{\partial \bfx} \end{array} \right] \sim \mathcal{N} 
\left( \bf{0}, \left[ \begin{array}{c c}  \kxxt & \partial \ksxt  \\ \partial \ksxt^\top &  \partial^2 \ksst  \end{array} \right] \right),
\label{eq:joint_differential}
\end{equation}
where %$\tilde{\bfK} = \bfK +\beta^{-1}\bfI$, and 
$\partial \kxst, \partial^2 \ksst$ are a matrices given by
\begin{eqnarray}
(\partial \ksxt)_{n,l} = \frac{\partial k(\bfx_n,\bfx_*)}{\partial x_*^{(l)}}, & \begin{array}{l} n=1, \cdots, \numData \\ l=1, \cdots, \latentDim \end{array} \label{eq:k_derivatives1}\\
(\partial^2 \ksst)_{i,l} = \frac{\partial^2 k(\bfx_*,\bfx_*)}{\partial x_*^{(i)} \partial x_*^{(l)}}. & \begin{array}{l} i=1, \cdots, \latentDim \\ l=1, \cdots, \latentDim  \end{array}
\label{eq:k_derivatives2}
\end{eqnarray}

The GP-LVM model provides an explicit mapping from the latent space to the observed space. This mapping defines the support of the observed data $\bfY$ as a $\latentDim$ dimensional manifold embedded into $\mathbb{R}^\dataDim$. If the covariance function of the model is continuous and differentiable, the Jacobian of the GP-LVM mapping is well-defined and the natural metric follows Eq.~\ref{eq:wishart}.
%it is possible to define the metric tensor according to \eqref{eq:metric}. 

It follows from Eq.~\ref{eq:joint_differential} and the properties of the GPs that the distribution of the Jacobian of the GP-LVM mapping is the product of $\dataDim$ independent Gaussian distributions (one for each dimension of the dataset) with mean $\bfmu_{J(j,:)}$ and covariance $\bfSigma_{J}$. For a every latent point $\bfx_*$ the Jacobian takes the following form:
\begin{align} \label{eq:J_distribution_GPLVM}
  &p(\bfJ_* \mid \bfY,\bfX, \bfx_*) = \prod_{j=1}^\dataDim \mathcal{N} (\bfJ_{j,:} \mid \bm{\mu}_{J_{j,:}}, \bfSigma_{J})=\\  
  &\prod_{j=1}^\dataDim\!\mathcal{N} (\bfJ_{j,:}\!\mid\!\partial\ksxt^\top\kxxt^{-1}\bfY_{:,j}\!, \partial^2 \ksst\!\!-\!\partial \ksxt^\top \kxxt^{-1} \partial \ksxt\!), \nonumber
\end{align}
which (c.f.\ Eq.~\ref{eq:wishart}) gives a distribution over the metric tensor $\bfG$ 
\begin{equation}
\bfG = \mathcal{W}_{\latentDim}( \dataDim, \bfSigma_J, \mathbb{E}[\bfJ^\top]\mathbb{E}[\bfJ]).
\end{equation}

From this distribution, the expected metric tensor can be computed as
\begin{equation}
\mathbb{E}[\bfJ^\top\bfJ] =  \mathbb{E}[\bfJ^\top]\mathbb{E}[\bfJ] + \dataDim \; \bfSigma_J.
\end{equation}

Note that the expectation of the metric tensor includes a covariance term. This implies that
the metric tensor expands as the uncertainty over the mapping increases. Hence, curve lengths
also increases when going through uncertain regions, and as a consequence geodesics will
tend to avoid these regions.
%This nice effect (displayed in Fig.~\ref{fig:cmu_MF} depends on the fact that in GP-LVM the covariance term depends on the values of the latent mapping.
%Not all the LVMs have this property; e.g.\ in GTM the expected metric tensor follows from \eqref{eq:gtm}.
%In the literature \citep{Bishop:gtm}, this quantity has been computed without taking into account the covariance term. In the case of GTM this omission makes no difference because the covariance of $\bfJ$ is constant, due to the fact that the covariance of the conditional distribution given by \ref{eq:gtm} does not depend on the latent space.

\begin{figure}[h]
  \includegraphics[trim=0 0 0 0,clip,scale=0.60]{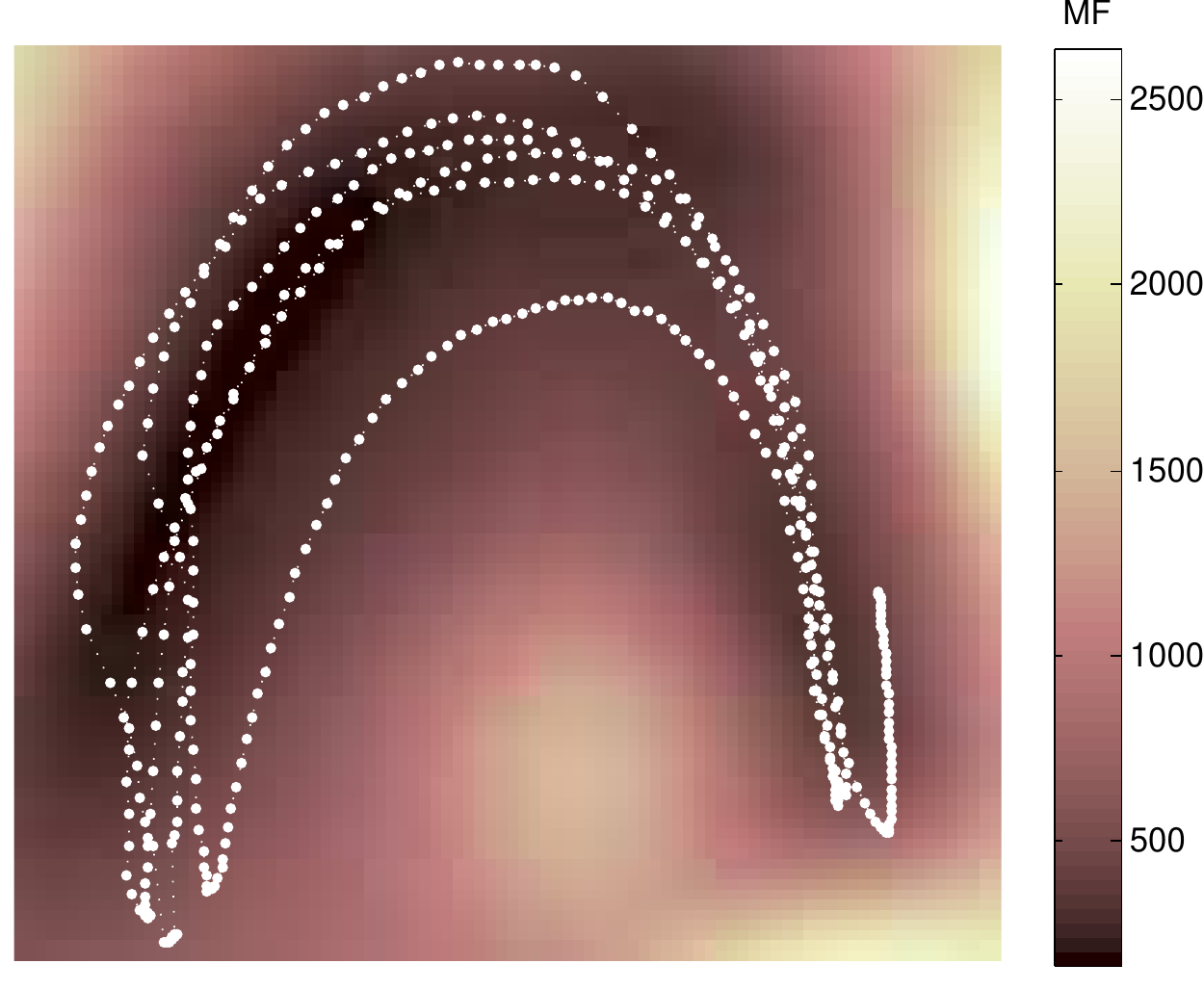}
  \vskip .5pc 
  \caption{GP-LVM latent space for the motion capture data (see section \ref{sec:exp} for details). White dots denote
    latent points $\bfx_n$ and the background colour is proportional to the magnification
    factor \eqref{eq:mag_fac}.}
  \label{fig:cmu_MF}
\end{figure}

The metric tensor defines the local geometric properties of the GP-LVM model and it can be used as a tool to data exploration. One way to visualise the tensor metric is through the differential volume of the high dimensional parallelepiped spanned by GP-LVM; this, for a latent dimension $\latentDim = 2$ is known as magnification factor and it has been introduced by \citep{Bishop:wsommag97} for generative topographic mapping (and self organising maps). Its explicit formulation for GP-LVM is given by
\begin{equation}
  \mathrm{MF} = \sqrt{\operatorname{det} \left( \mathbb{E}[\bfJ^\top\bfJ] \right)}.
  \label{eq:mag_fac}
\end{equation}
An example of the magnification factor is shown in Fig.~\ref{fig:cmu_MF}.

\subsection{COMPUTING GEODESICS}\label{sec:comp_geo}

Given a latent space endowed with an expected Riemannian metric, we
now consider how to compute geodesics (shortest paths) between given
points. Once a geodesic is computed its length can be evaluated
through numerical integration of Eq.~\ref{eq:curve_length}.

The obvious solution to the shortest path problem is to discretise the
latent space and compute shortest paths on the resulting graph using
e.g.\ Dijkstra's algorithm \citep{Cormen:book90}. The computational
complexity of this approach, however, grows exponentially with the
dimensionality of the latent space and the approach quickly becomes
infeasible. Further, this approach will also introduce discretisation
errors due to the finite size of the graph.

Instead we solve the geodesic differential equation
\eqref{eq:geodesic_ode} numerically.  This scales more gracefully as
it only involves a discretisation of the geodesic curve which is
always one-dimensional independently of the dimension of the latent
space. The 2nd order ODE in \eqref{eq:geodesic_ode} can be rewritten in a 
standard way as a system of 1st order ODE’s, which we can solve 
using a four-stage implicit Runge-Kutta 
method\citep{bvp4c}\footnote{We use an off-the-shelf numerical solver
(\textsf{bvp5c} in Matlab$^{\circledR}$); runnig times and computational 
cost are provided in the reference.}. This gives a smooth solution which
is fifth order accurate. 
%See \citep{hauberg:nips:2012} for more details about the algorithm. 
Alternatively, such equations can be solved by repeated Gaussian 
process regression \citep{hennig:aistats:2014}.

To evaluate Eq.~\ref{eq:geodesic_ode} we need the derivative of the
expected metric:
\begin{equation}
\frac{\partial\vect\mathbb{E}[\bfG(\bfx)]}{\partial \bfx} = \frac{\partial\vect(\mathbb{E}[\bfJ^\top]\mathbb{E}[\bfJ] + \dataDim \cdot \cov(\bfJ,\bfJ))}{\partial \bfx}.
\end{equation}
For the GP-LVM this reduces to computing the derivatives of
the covariance function $k$.  Given two vectors $\bfx_1, \bfx_2 \in \mathbb{R}^{\latentDim}$, 
a widely used covariance function is the \emph{squared exponential}
(or \emph{RBF}) kernel
\begin{equation}
k(\bfx_1,\bfx_2)= \alpha \, \exp \left( -\frac{\omega}{2} \parallel \bfx_1 - \bfx_2 \parallel_2^2 \right).
\end{equation}
We choose here the \emph{RBF} as an illustrative example, but our 
approach apply to any other kernel that leads to a differential mapping.
This function is differentiable in $\bfx$ and will be used 
here (and in Section~\ref{sec:exp}) to provide a specific algorithm. 
We explicitly compute Eq.~\ref{eq:k_derivatives1} and \ref{eq:k_derivatives2} 
for the squared exponential kernel to have an explicit form of Eq.~\ref{eq:J_distribution_GPLVM}:
\begin{align}
 &\left(\partial \tilde{\bfK}_{\bfx_*,\bfx}\right)_{1,j} \:\: = \: - \omega (x_*^{(j)} - x^{(j)}) \, k(\bfx_*, \bfx) \\
 &\left(\partial^2 \tilde{\bfK}_{\bfx_1,\bfx_2}\right)_{i,l} = \\
 &= \left\{ \begin{array}{ll} 
  \omega (x_1^{(i)} - x_2^{(i)}) (x_1^{(l)} - x_2^{(l)}) \, k(\bfx_1,\bfx_2), & i \neq l \\  
  \omega (\omega(x_1^{(i)} - x_2^{(i)})^2-1) \, k(\bfx_1,\bfx_2), & i = l
\end{array} \right. \notag
\end{align}

Due to symmetry, the upper triangular of the Hessian matrix is sufficient to the computation. Note that, for our choice of kernel, the Hessian is diagonal and constant for $\bfx_1 = \bfx_2$, which is the case of $\tilde{\partial^2\kss}$, so there is no need to compute its derivative (which appears in the expression of $\partial\vect\bfG$). 

\section{EXPERIMENTS AND RESULTS}\label{sec:exp}
\begin{figure}[t]
\begin{center}
  \fboxsep=0mm
  \fbox{\includegraphics[width=1.45cm]{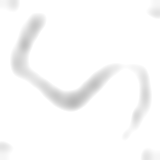}} %&
  \fbox{\includegraphics[width=1.45cm]{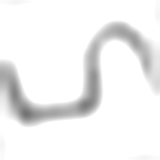}} %&
  \fbox{\includegraphics[width=1.45cm]{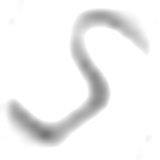}} %&
  \fbox{\includegraphics[width=1.45cm]{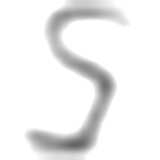}} %&
  \fbox{\includegraphics[width=1.45cm]{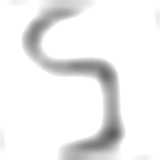}} \\
  \fbox{\includegraphics[width=1.45cm]{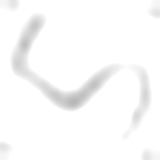}} %&
  \fbox{\includegraphics[width=1.45cm]{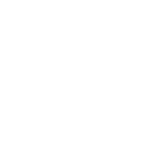}} %&
  \fbox{\includegraphics[width=1.45cm]{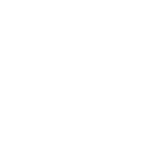}} % &
  \fbox{\includegraphics[width=1.45cm]{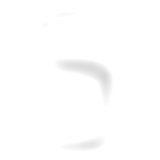}} %&
  \fbox{\includegraphics[width=1.45cm]{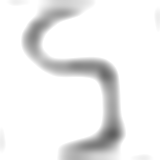}} %\\
\end{center}
\vskip .5pc 
\caption{Rotated digit. Inference after sampling over the latent space following
  the Geodesic distance (top row) and the Euclidean distance (bottom row); see also
  Fig.~\ref{fig:introduction}. Images are inverted and bicubically upscaled for improved viewing.}
\label{fig:digit_sequence}
\end{figure}

  Section~\ref{sec:intro} shows a first motivating example:
  a single image of a hand-written digit is rotated from 0 to 360 degrees to produce
  200 rotated images. We then estimate\footnote{Software from the Machine Learning 
  group, University of Sheffield http://staffwww.dcs.shef.ac.uk/people
  /N.Lawrence/software.html}
  a GP-LVM model with a $\latentDim = 2$ dimensional
  latent space; the latent space is shown in Fig.~\ref{fig:introduction}. We interpolate
  two points using either a straight line or a geodesic, and reconstruct images along
  these paths. The results in Fig.~\ref{fig:digit_sequence}
  show the poor reconstruction of the straight-line interpolator. The core
  problem with this interpolator is that it goes through regions with little data support,
  meaning the resulting reconstruction will be similar to the average of the entire
  data set.
  
  In the next two sections we consider experiments on real data, but our
  results are similar to the synthetic digit experiment. First, we consider
  images of rotating objects (Section~\ref{sec:coil}), and then motion capture data
  (Section~\ref{sec:mocap}).

\subsection{IMAGES OF ROTATING OBJECTS}\label{sec:coil}
  We consider images from the COIL data set \citep{coil}, which consists of images
  from a fixed camera depicting 100 different objects on a motorised turntable
  against a black background. Each image is acquired after a 5 degree rotation
  of the turntable, giving a total of 72 images per object. Here we consider
  the images of object $74$ (a rubber duck), but similar results are attained
  for other objects.
  
\begin{figure}[t]
\begin{center}
  \fboxsep=0mm
  {\includegraphics[width=1.45cm]{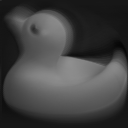}} %&
  {\includegraphics[width=1.45cm]{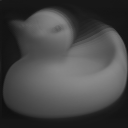}} %&
  {\includegraphics[width=1.45cm]{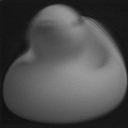}} %&
  {\includegraphics[width=1.45cm]{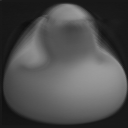}} %&
  {\includegraphics[width=1.45cm]{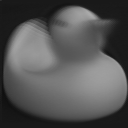}} \\
  {\includegraphics[width=1.45cm]{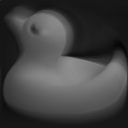}} %&
  {\includegraphics[width=1.45cm]{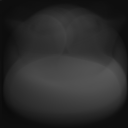}} %&
  {\includegraphics[width=1.45cm]{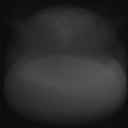}} % &
  {\includegraphics[width=1.45cm]{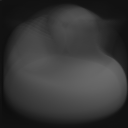}} %&
  {\includegraphics[width=1.45cm]{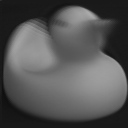}} %\\
\end{center}
\vskip .5pc 
\caption{COIL image reconstruction. Inference after sampling over the latent space following
  the geodesic (top row) and the Euclidean straight line (bottom row).}
\label{fig:duck1}
\end{figure}

  We estimate a $\latentDim = 2$ dimensional latent space using GP-LVM, and interpolate
  two latent points using either a straight line or a geodesic. Reconstructed images
  along the interpolated paths are shown in Fig.~\ref{fig:duck1}. It is clear that the 
  geodesic gives a better interpolation as it avoids regions with high uncertainty.
  
\begin{figure}[t]
\begin{center}
  \includegraphics[width=0.45\textwidth]{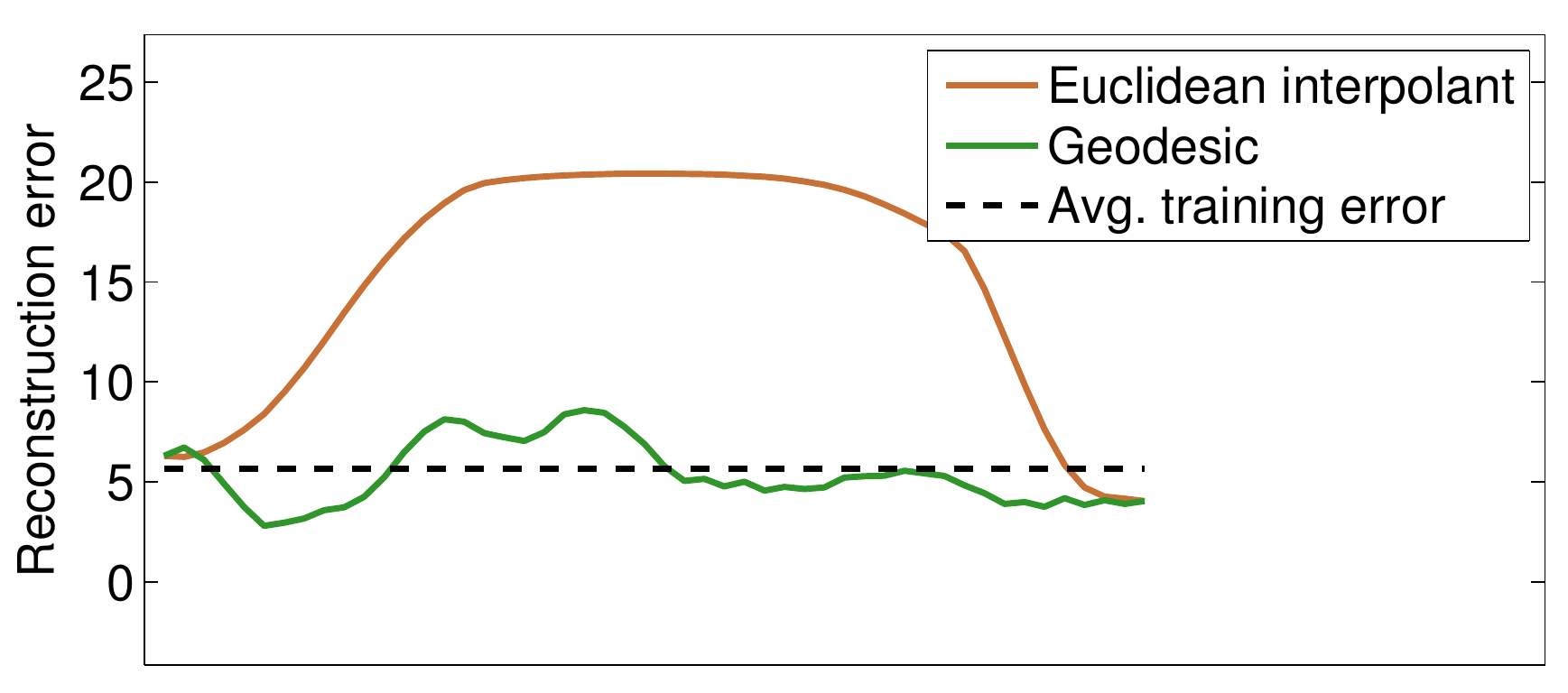}
\end{center}

\vskip .5pc 
\caption{COIL reconstruction error. Inference after sampling over the latent space following
  the geodesic (green) and the Euclidean straight line (brown). For reference, the average
  reconstruction error of the latent observations is shown as well (dashed).
  This figure is best viewed in colour.}
\label{fig:duck2}
\end{figure}

  To measure the quality of the different interpolators we reconstruct 50 images
  equidistantly along each interpolating path and measure the distance to the
  nearest neighbour in the training data. This is shown in Fig.~\ref{fig:duck2},
  which, for reference, also shows the average reconstruction error of the latent representations
  of the training data,
  \begin{align}
    \textrm{Avg. training error} = \frac{1}{N}\sum_{n=1}^N \| \mathbb{E}\left[f(\bfx_n)\right] - \bfy_n \|.
  \end{align}
  It is clear that the straight line interpolator
  performs poorly away from the end-points, while the geodesic provides errors which
  are comparable to the average error of the latent representation of the training data.

\subsection{HUMAN MOTION CAPTURE}\label{sec:mocap}
\begin{figure}[t]
  \begin{center}
  \begin{tabular}{c c c c c}
    \includegraphics[trim=0 0 0 0,clip,scale=0.25]{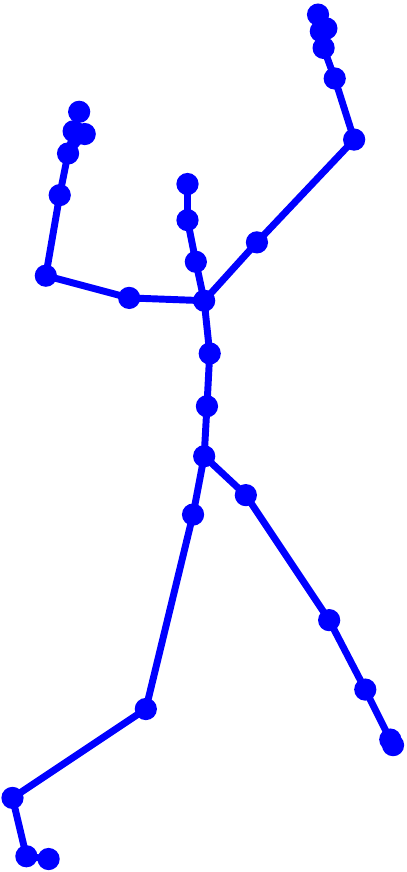}  &
    \includegraphics[trim=0 0 0 0,clip,scale=0.25]{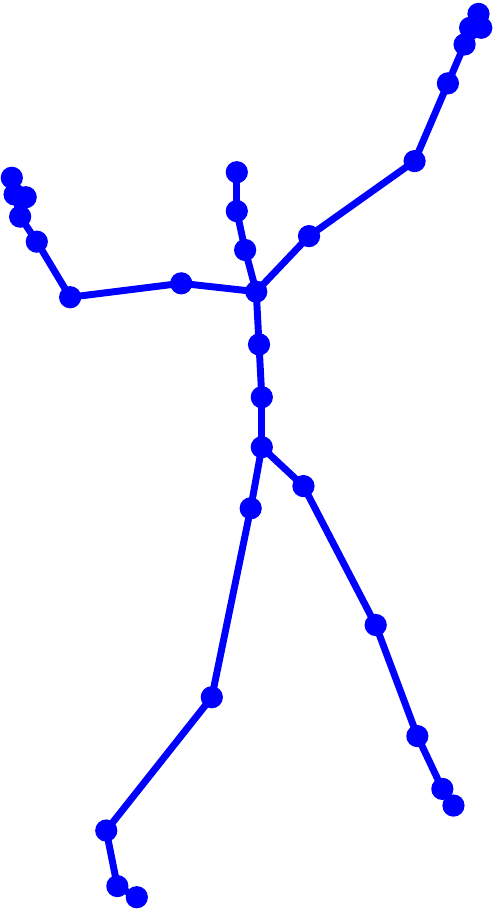}  &
    \includegraphics[trim=5 0 0 0,clip,scale=0.25]{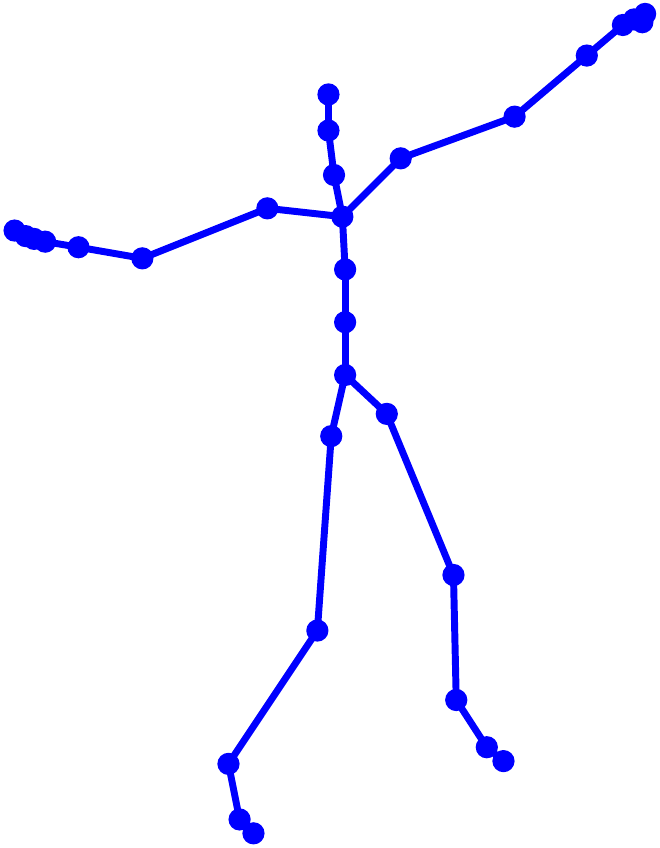}  &
    \includegraphics[trim=10 0 0 0,clip,scale=0.25]{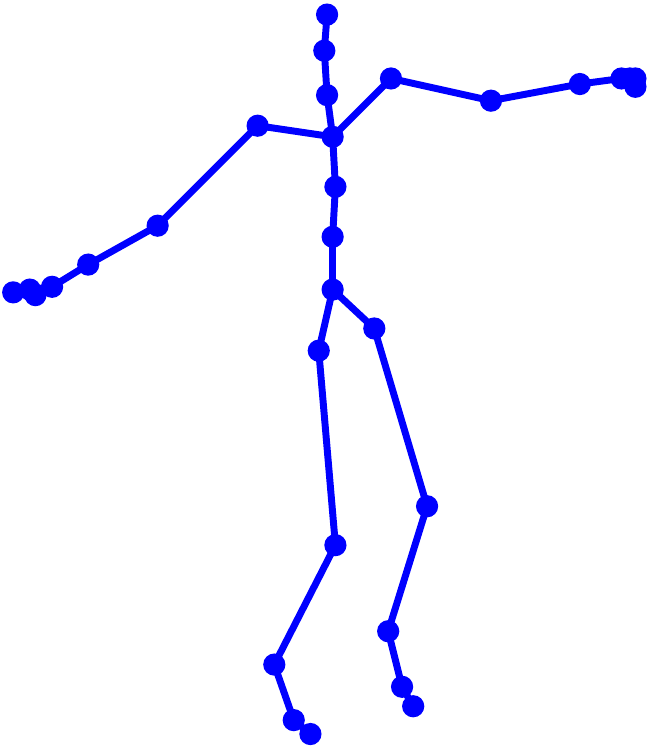} &
    \includegraphics[trim=0 0 0 0,clip,scale=0.25]{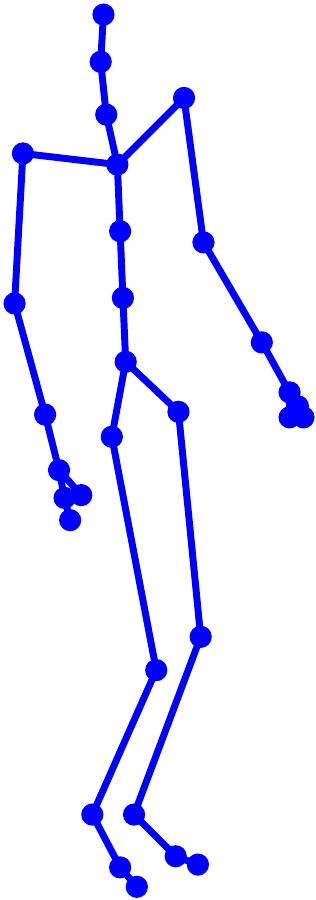}
  \end{tabular}
  \end{center}
  \vskip .5pc 
  \caption{Example poses from the motion capture data. These poses are temporarly
    between the end-points of the interpolating curves, i.e.\ they are comparable
    to the interpolated reconstructions.}
  \label{fig:seq_training}
  \begin{center}
  \begin{tabular}{c c c c c}
    \includegraphics[trim=0 0 0 0,clip,scale=0.25]{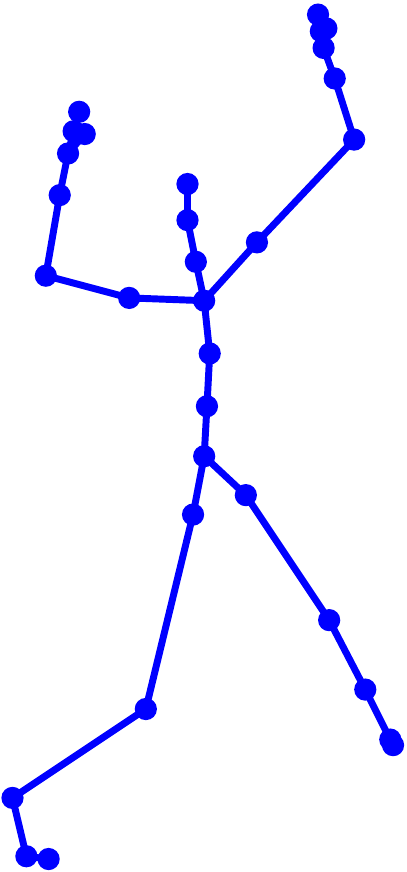}   &
    \includegraphics[trim=-10 0 0 0,clip,scale=0.25]{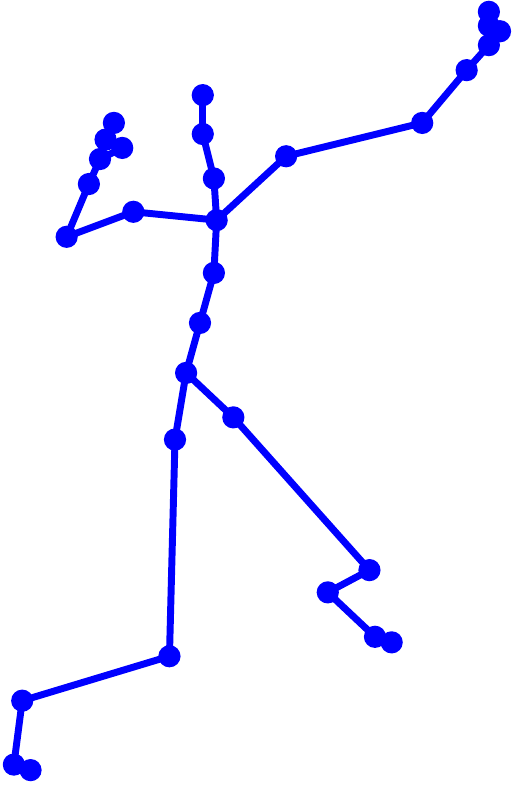} &
    \includegraphics[trim=-10 0 0 0,clip,scale=0.25]{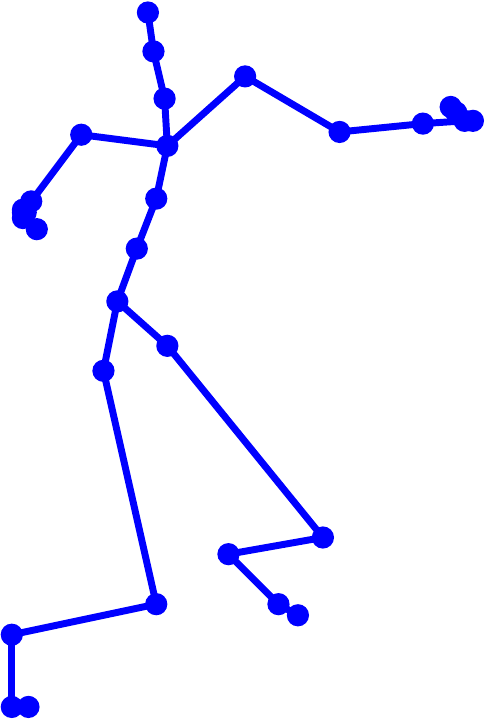} &
    \includegraphics[trim=-15 0 0 0,clip,scale=0.25]{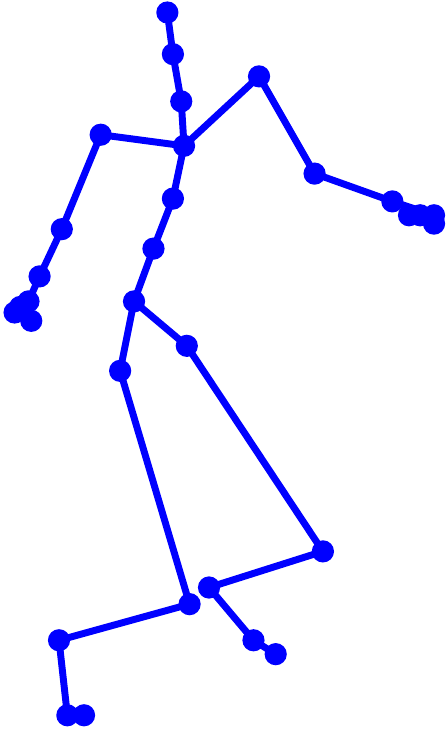} &
    \includegraphics[trim=-20 0 0 0,clip,scale=0.25]{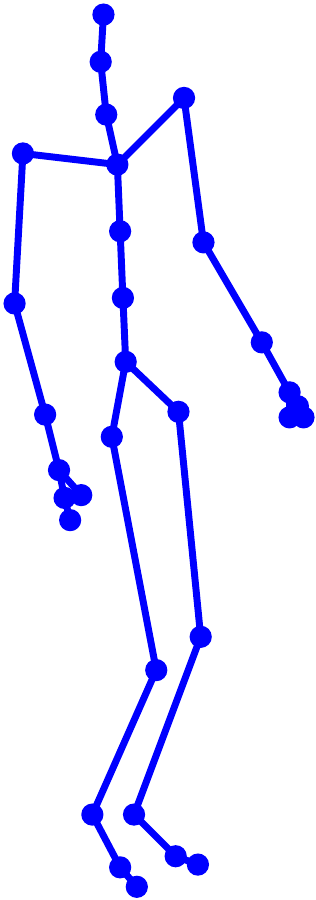}
  \end{tabular}
  \end{center}
  \vskip .5pc 
  \caption{Interpolated poses according to the straight-line interpolant. In particular,
    note the bending of the knees and the retraction of the arms, which do not occur 
    in the training data.}
  \label{fig:seq_euclid}
  \begin{center}
  \begin{tabular}{c c c c c}
    \includegraphics[trim=0 0 0 0,clip,scale=0.25]{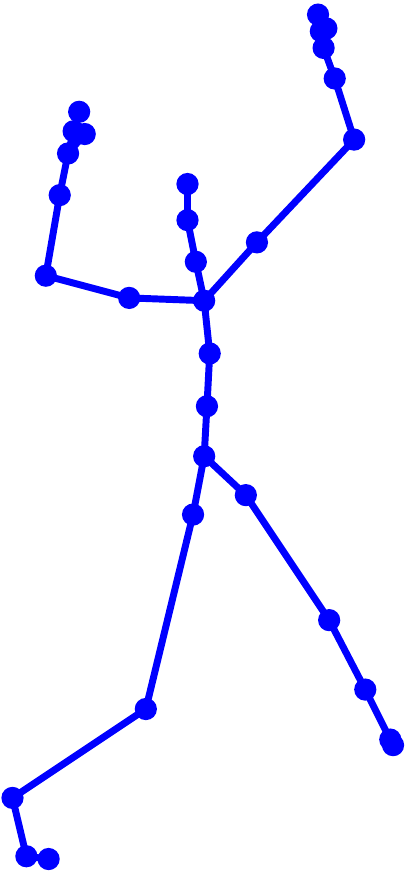} &
    \includegraphics[trim=0 0 0 0,clip,scale=0.25]{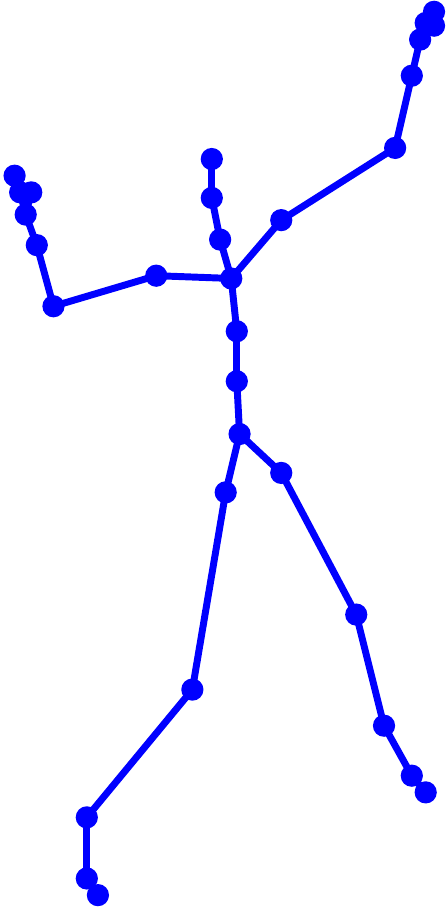} &
    \includegraphics[trim=0 0 0 0,clip,scale=0.25]{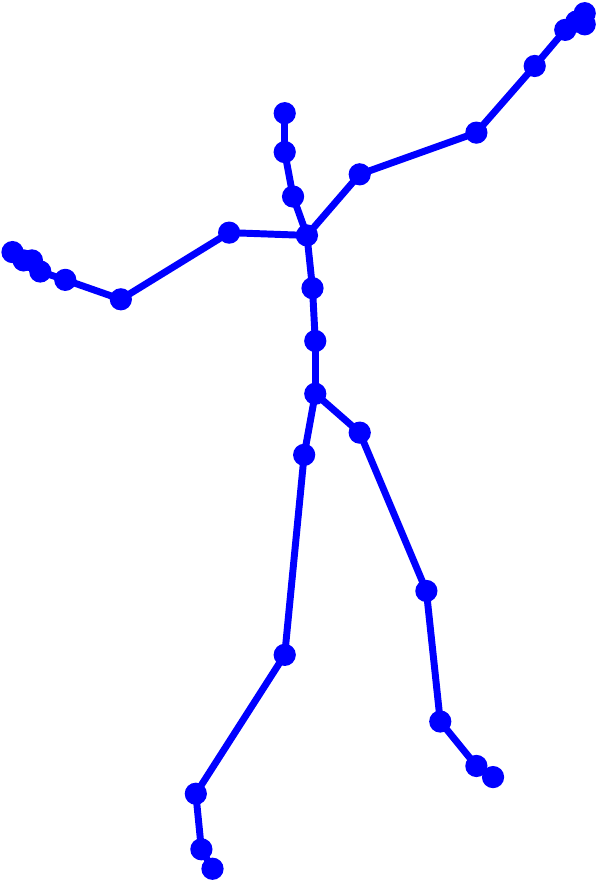} &
    \includegraphics[trim=0 0 0 0,clip,scale=0.25]{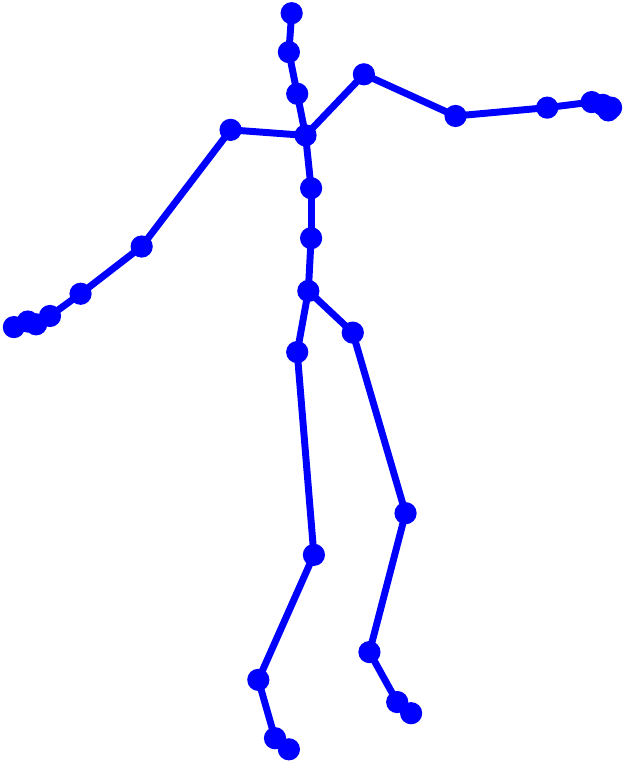} &
    \includegraphics[trim=0 0 0 0,clip,scale=0.25]{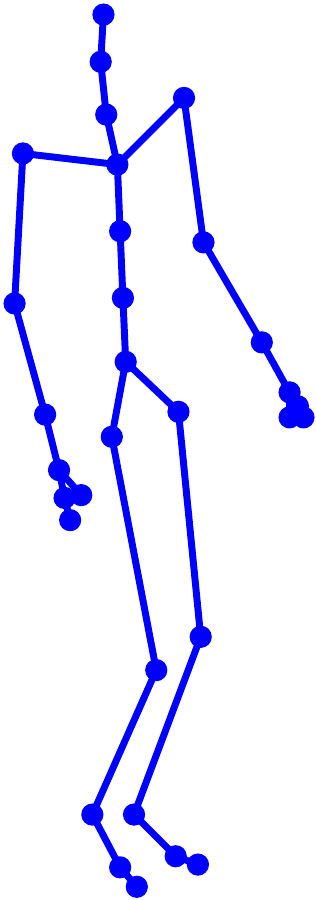}
  \end{tabular}
  \end{center}
  \vskip .5pc 
  \caption{Interpolated poses according to the geodesic. These are visually similar
    to the poses in Fig.~\ref{fig:seq_training}.}
  \label{fig:seq_geodesic}
\end{figure}

  We next consider human motion capture data from the \emph{CMU Motion Capture
  Database}\footnote{\url{http://mocap.cs.cmu.edu/}}. Specifically, we study
  motion 16 from subject 22, which is a repetitive \emph{jumping jack} motion.
  Each time instance of this data set consists of a human pose as acquired by a
  marker-based motion capture system; see Fig.~\ref{fig:seq_training} for example
  data. We represent each pose by the three-dimensional joint positions, i.e. as
  a vector $\bfy_{n,:} \in \mathbb{R}^{3P}$, where $P$ denotes number of joint positions.

\begin{figure}[h]
  \includegraphics[trim=0 0 0 0,clip,scale=0.58]{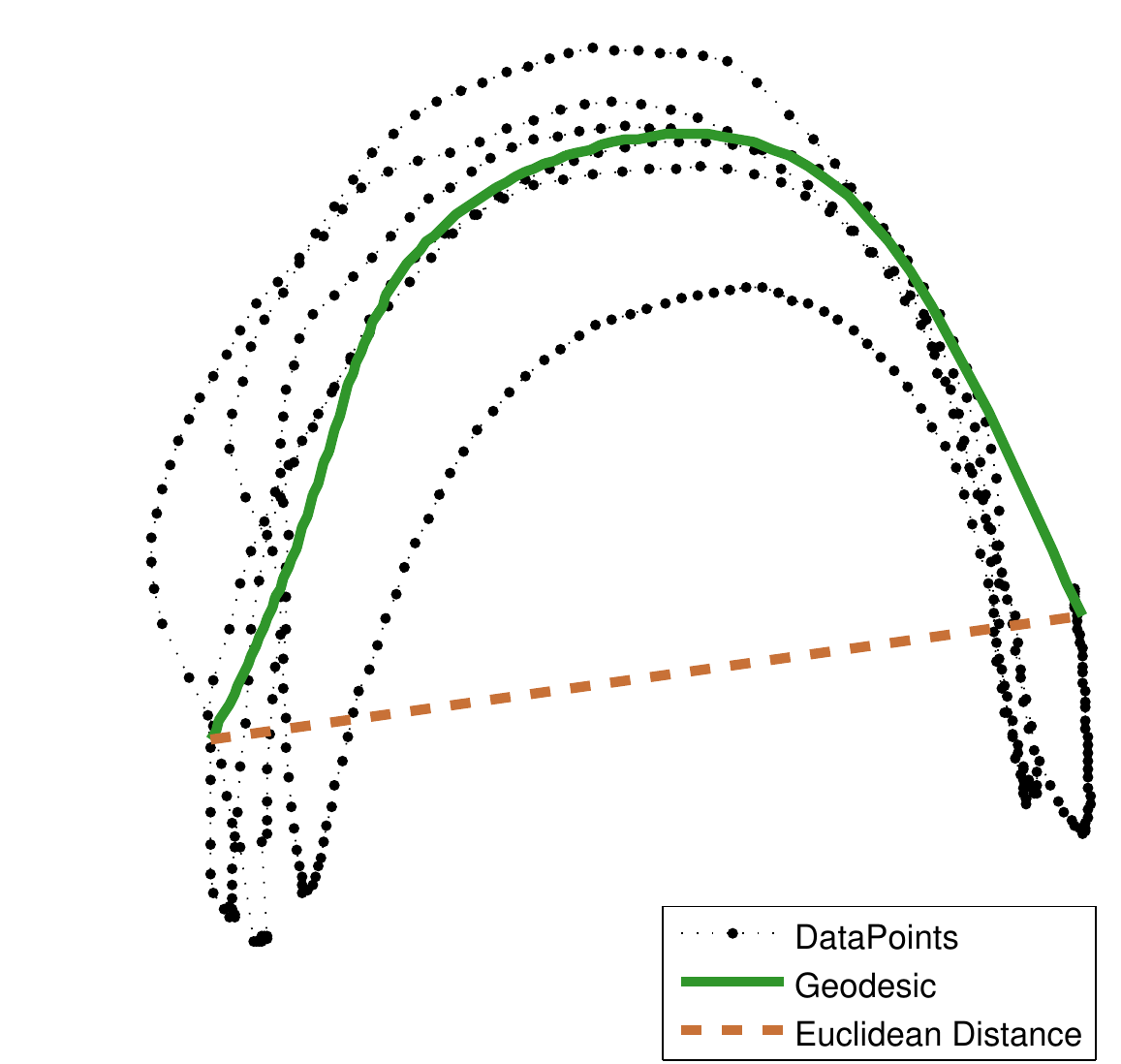}
 % \vspace{1in} 
  \vskip .5pc   
  \caption{Latent space for the \textit{jumping jacks} motion capture data, trained with 
  GP-LVM. Black dots denote latent points $\bfx_n$. The green curve denotes the geodesic 
  interpolant, while the dashed brown curve is the straight-line interpolant.
  \vspace{0.1cm}}
  \label{fig:cmu_geodesic_metric}
\end{figure}

  We estimate a GP-LVM using dynamics \citep{Damianou:vgpds11} as is common for this type of data \citep{wang:pami:2008}.
  The resulting latent space is shown in Fig.~\ref{fig:cmu_geodesic_metric}, and the
  metric tensor is shown in Fig.~\ref{fig:cmu_MF}.
  %, where the  background colour is proportional to the magnification factor \ref{eq:mag_fac} of the
  %Riemannian metric.
  As can be seen, the latent points $\bfx_{n,:}$ follow a periodic
  pattern as expected for this motion, and the metric tensor is generally smaller in regions
  of high data density.
  
  We pick two latent extremal points of the motion ($\bfx_1$ and $\bfx_T$) and interpolate 
  them using the Euclidean straight line and the expected Riemannian geodesic.
  Fig.~\ref{fig:cmu_geodesic_metric} shows the interpolants: again, the geodesic
  follow the trend of the data while the straight line goes through regions with
  high model uncertainty. Reconstructed poses along the interpolants are shown
  in Fig.~\ref{fig:seq_euclid} and \ref{fig:seq_geodesic}. A comparison with the
  intermediate poses ($\bfx_2 \ldots \bfx_{T-1}$) in the training sequence (see
  Fig.~\ref{fig:seq_training}) reveals that the geodesic interpolant is a more truthful
  reconstruction compared to that of the straight line.
  
\begin{figure}[h]
  \includegraphics[trim=0 0 0 0,clip,scale=0.6]{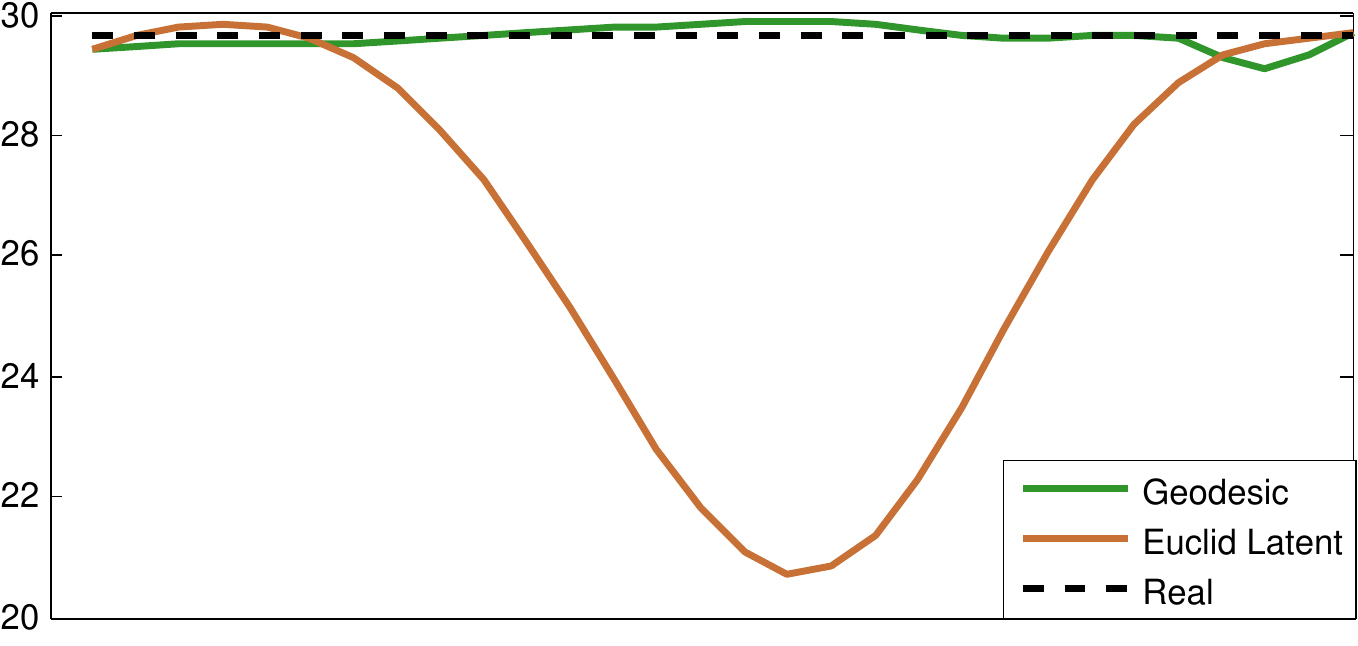}
  \vskip .5pc 
  \caption{Length, in centimetres, of the subjects forearm during latent space interpolation.
    The green curve is defined according to the geodesic interpolant, and the brown dashed
    curve according to the straight-line interpolant. For reference, the black dashed line
    shows the true length.}
  \label{fig:cmu_len}
\end{figure}

  To measure the quality of the reconstruction we note that the length of the subject's
  limbs should stay constant throughout the sequence. Our representation does, however,
  not enforce this constraint. Fig.~\ref{fig:cmu_len} shows the length of the subjects
  forearm for the two reconstructions along with the correct length. The straight line
  interpolant drastically changes the limb lengths, while the geodesic matches the ground
  truth well. Similar observations have been made for other limbs.

\section{DISCUSSION AND FUTURE WORK}\label{sec:discussion}
  When the mapping between a latent space and the observation space is not isometric
  (the common case for nonlinear mappings), a Euclidean distance measure in the latent
  space does not match that of the original observation space. In fact, the distance measures
  in the latent and observation spaces can be arbitrarily different. This makes it 
  difficult to perform any meaningful statistical operation directly in the latent space
  as the used metric is difficult to interpret.
  
  We solve this issue by carrying the metric from the observation space into the latent
  space in the form of a \emph{random Riemannian metric}. This gives a distribution
  over a smoothly changing local metric at each point in the latent space. We then provide
  an expression for the \emph{expected} local metric and show how shortest paths (geodesics)
  can be computed numerically under the resulting metric. These geodesics provide natural
  generalisations of straight-lines and are, thus, suitable for interpolation under the
  new metric.
  
  For the GP-LVM model the expected metric depends on the uncertainty of the model, such
  that distances become longer in regions of high uncertainty. This effectively forces
  geodesic curves to avoid uncertain regions in the latent space, which is the desired
  behaviour for most applications. It is worth noting that a similar analysis for the
  GTM does \emph{not} provide a metric with this capacity as the uncertainty is constant
  in this model.
  
  The idea of considering the expected metric is practical as it turns the latent
  space into a Riemannian manifold. This opens up to many applications as statistical
  operations are reasonably well-understood in these spaces. E.g.\ tracking can be performed
  in the latent space through a Riemannian Kalman filter \citep{hauberg:jmiv:2013},
  classification can be done using the geodesic distance, etc.
  
  It is, however, potentially misleading to only consider the expectation of the metric
  rather than the entire distributions of metrics. Although, if the latent dimension is 
  much lower than the data dimension, it can be shown that the distribution of the metric
  concentrates around its mean. But in general \emph{random
  Riemannian manifolds} are mathematically less well-understood, e.g.\ it is known
  that geodesics are almost surely not length minimising curves under a random metric
  \citep{legatta:2014}. We are suggesting that manifolds derived from data are necessarily 
  uncertain, and there is much to gain from further consideration of these spaces, which 
  then naturally lead to distributions over geodesics, distances, angles, curvature and 
  so forth.
  
  In this paper we have only considered how geometry can be used to understand an
  already estimated LVM, but it is also worth considering if this geometry can be used
  when estimating the LVM. E.g.\ it is worth investigating if a prior on the curvature
  of the latent manifold is an effective way to influence learning.

%%%%%%%%%%%%%%%%%%%%%%%
 
%\newpage 

\subsubsection*{Acknowledgements} 
 
The authors found great inspiration in the discussions at the
\emph{1st Braitenberg Round table} on \emph{Probabilistic numerics
and Random Geometries}. This research was partially funded by European 
research project EU FP7-ICT (Project Ref 612139 "WYSIWYD) 
and by Spanish research project TIN2012-31377. 
S.H.~is funded in part by the Danish Council for Independent Research 
(Natural Sciences); the Villum Foundation; and an 
\texttt{amazon.com} machine learning in education award.

\newpage 

\subsubsection*{References} 
%\vskip .5pc 
\vspace{-7mm}
\bibliographystyle{abbrvnat}
\bibliography{refUAI,lawrence,other,zbooks}

\end{document}